\documentclass{article}
\usepackage{ijcai21}

\usepackage{times}
\usepackage{soul}
\usepackage{url}
\usepackage[hidelinks]{hyperref}
\usepackage[utf8]{inputenc}
\usepackage[small]{caption}
\usepackage{graphicx}
\usepackage{amsmath}
\usepackage{amsthm}
\usepackage{booktabs}
\usepackage{algorithm}
\usepackage{algorithmic}

\usepackage{setspace}
\usepackage{amssymb} 
\usepackage[labelformat=empty, font=footnotesize]{subcaption}
\usepackage{mathtools}
\usepackage{mathrsfs}
\usepackage{bm}

\usepackage{tikz}
\usepackage{atbegshi,picture}
\newcommand{\myleftstd}{1in}
\newcommand\copyrighttext{%
  \footnotesize \textcopyright 2021 International Joint Conference on Artificial Intelligence (IJCAI) \textcolor{blue}{\url{https://www.ijcai-21.org}} \newline Abawi, F., Weber, T., \& Wermter, S. (2021). GASP: Gated Attention for Saliency Prediction. \textit{In Proceedings of the Thirtieth International Joint Conference on Artificial Intelligence, IJCAI-21} (pp. 584–591). International Joint Conferences on Artificial Intelligence Organization.
  \providecommand{\doi}{\href{https://doi.org/10.24963/ijcai.2021/81}{10.24963/ijcai.2021/81}}
  DOI: \doi
  }
\ifthenelse{\isodd{\thepage}}
   {\newcommand{\myleftmargin}{\oddsidemargin+\myleftstd}}
   {\newcommand{\myleftmargin}{\evensidemargin+\myleftstd}}
\AtBeginShipoutNext{\AtBeginShipoutUpperLeft{%
  \put(\dimexpr\myleftmargin\relax,-1.2cm){\parbox{\textwidth}{\normalsize\sffamily\noindent{\centering  \hfill}}}%
  \begin{tikzpicture}[remember picture,overlay]
  \node[anchor=south,yshift=10pt] at (current page.south) {\fbox{\parbox{\dimexpr\textwidth-\fboxsep-\fboxrule\relax}{\copyrighttext}}};
  \end{tikzpicture}%
}}

\title{GASP: Gated Attention for Saliency Prediction}

\author{
Fares Abawi\footnote{Contact Author}\and
Tom Weber\And
Stefan Wermter\\
\affiliations
Knowledge Technology, University of Hamburg, Germany\\
\emails
\{abawi, tomweber, wermter\}@informatik.uni-hamburg.de
}

\begin{document}

\maketitle

\begin{abstract}
Saliency prediction refers to the computational task of modeling overt attention. Social cues greatly influence our attention, consequently altering our eye movements and behavior. 
To emphasize the efficacy of such features, we present a neural model for integrating social cues and weighting their influences. Our model consists of two stages. During the first stage, we detect two social cues by following gaze, estimating gaze direction, and recognizing affect. These features are then transformed into spatiotemporal maps through image processing operations. The transformed representations are propagated to the second stage (\textit{GASP}) where we explore various techniques of late fusion for integrating social cues and introduce two sub-networks for directing attention to relevant stimuli. Our experiments indicate that fusion approaches achieve better results for \textit{static} integration methods, whereas non-fusion approaches for which the influence of each modality is unknown, result in better outcomes when coupled with recurrent models for \textit{dynamic} saliency prediction. We show that gaze direction and affective representations contribute a prediction to ground-truth correspondence improvement of at least 5\% compared to dynamic saliency models without social cues. Furthermore, affective representations improve GASP, supporting the necessity of considering \textit{affect-biased attention} in predicting saliency.\footnote{Code:  \url{http://software.knowledge-technology.info\#gasp}}
\end{abstract}

\begin{figure}[t!]
\centering
\includegraphics[width=0.9\linewidth]{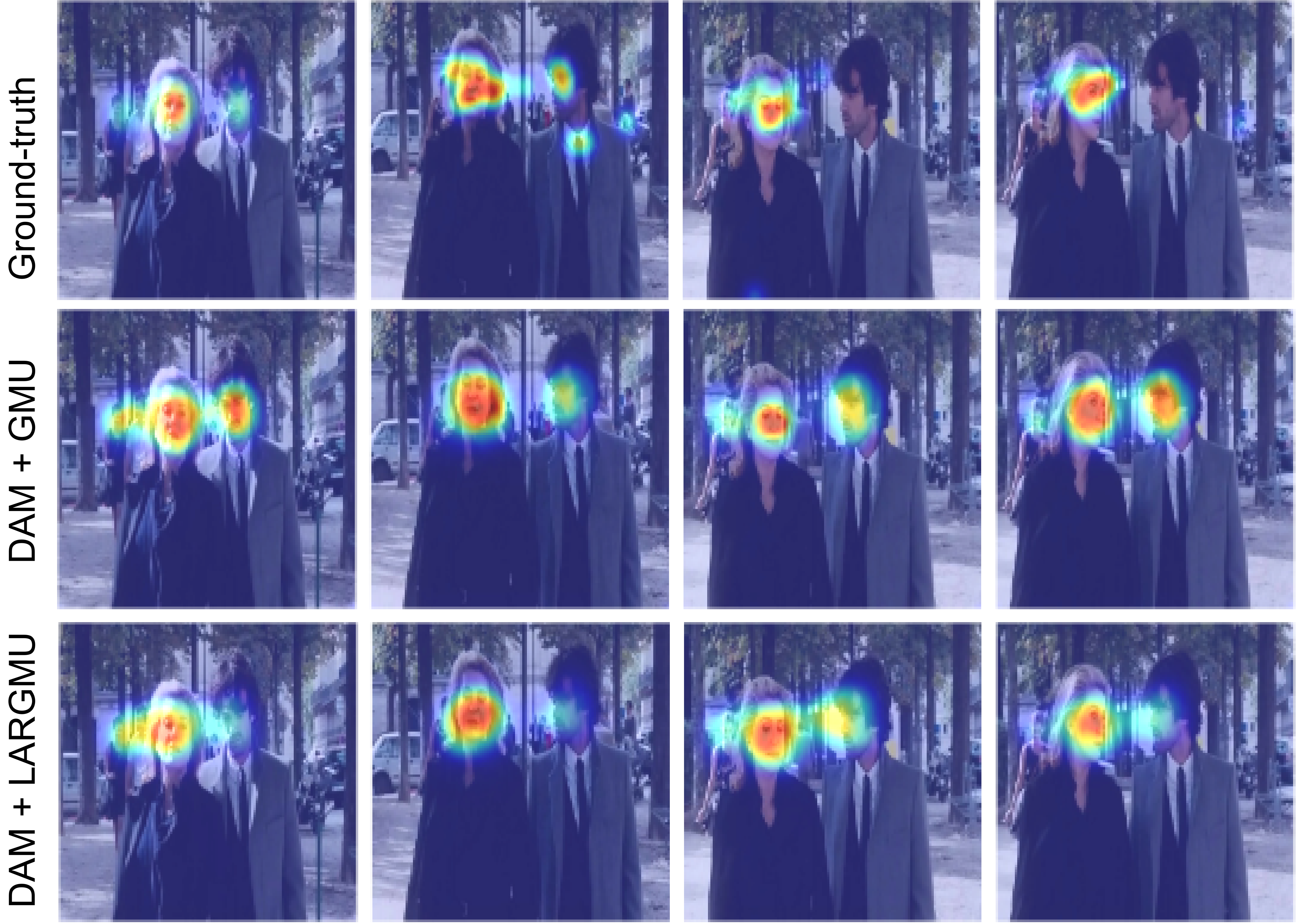}
\caption{Frame predictions on the Coutrot Database 1~\protect\cite{coutrot2014audiovisual}. DAM + GMU: Directed Attention Module followed by the Gated Multimodal Unit for static integration (middle); DAM + LARGMU: Directed Attention Module followed by the Late Attentive Recurrent GMU for sequential integration (bottom).}
\label{fig:samples}
\end{figure}

\section{Introduction}

Attending to regions or objects in our perceptual field implies an interest in acting towards them. Humans convey their attention by fixating their eyes upon those regions. By modeling fixation, we gain an understanding of the events that attract attention. These attractors are represented in the form of a \textit{Fixation Density Map} (FDM),
displaying blurred peaks on a two-dimensional map, centered on the eye \textit{Fixation Point} (FP) of each individual viewing a frame. The FDM is a visual representation of saliency, a useful indicator of what attracts human attention. 

Early computational research focused on bottom-up saliency, by which the conspicuity of regions in the visual field was purely dependent on the stimuli~\cite{itti2001computational,bruce2009saliency}. On the other hand, task-driven approaches are top-down models utilizing supervised learning for performing tasks and allocating attention to regions or objects of interest. Combining face detections with low-level features 
has been shown to outperform bottom-up saliency models agnostic to social entities in a scene. Birmingham \textit{et al.}~\shortcite{birmingham2009saliency} corroborate the advantage of facial features in modeling saliency. They establish that when social stimuli are present, humans tend to fixate on facial features, a phenomenon weakly portrayed by bottom-up saliency detectors. 
Moreover, studies on human eye movements indicate that bottom-up guidance is not strongly correlated with fixation, which is rather influenced by the task~\cite{foulsham2011modeling}. The existence of social stimuli in a scene alters fixation patterns, supporting the notion that even with the lack of an explicit task, we form intrinsic goals for guiding our gaze.

Although facial features attract attention, studies show that humans tend to follow the gaze of observed individuals~\cite{bylinskii2016should}.
Additionally, psychological studies~\cite{pritsch2017perception} 
indicate a preference in attending towards emotionally salient stimuli over neutral expressions, a phenomenon described as \textit{affect-biased attention}. 
By augmenting saliency maps with emotion intensities, affect-biased saliency models show significant improvement over affect-agnostic models~\cite{fan2018emotional,cordel2019emotion}. These approaches, although exclusive to static saliency models, are not limited to facial expressions, allowing for a greater domain coverage irrespective of the presence of social entities in a scene. 

In light of the social stimuli relevance to modeling attention, we design a model to predict the FDM of multiple human observers watching social videos as shown in~\autoref{fig:samples}. Such models employ top-down and bottom-up strategies operating on a sequence of images, a task referred to as \textit{dynamic saliency prediction}~\cite{bak2017spatio,borji2019saliency}. Our model utilizes multiple social cue detectors, namely gaze following and direction estimation, as well as facial expression recognition.  We integrate the eye gaze and affective social cues, each with its spatiotemporal representation, as input to our saliency prediction model. We describe the resulting output from each social cue detector as a \textit{feature map} (FM). We also introduce a novel FM weighting module, assigning different intensities to each FM in a competitive manner representing its priority. Each representation is best described as a \textit{priority map} (PM), combining top-down and bottom-up features to prioritize regions that are most likely to be attended. We refer to the final model output as the Predicted FDM (PFDM).

\begin{figure*}[ht]
\centering
\includegraphics[width=1.0\linewidth]{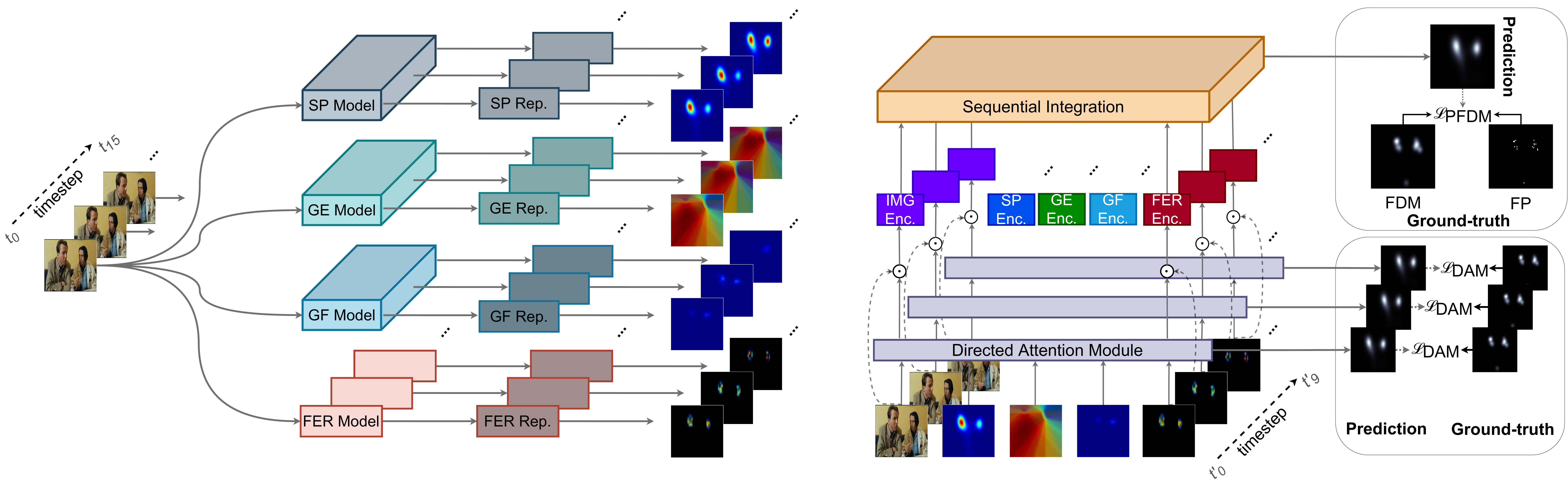}
\caption{Overview of our sequential two-stage model. SCD (left) extracts and transforms social cue features to spatiotemporal representations. GASP (right) acquires the representations and integrates features from the different modalities.} 
\label{fig:full_model}
\end{figure*}

Our work is motivated by the following findings:
\textbf{1)} Task-driven strategies are pertinent to predicting saliency~\cite{foulsham2011modeling};
\textbf{2)} Changes in motion contribute to the relevance of an object, underlining the importance of spatiotemporal features for predicting saliency~\cite{min2020multimodal};
\textbf{3)} Psychological studies indicate that attention is driven by social stimuli~\cite{salley2016conceptualizing}.
To address the first finding, we state that our approach is task-driven by virtue of supervision since the objective is predicated on modeling multiple observer fixations. Although the datasets employed in this study were collected under a free-viewing condition, the top-down property is arguably maintained due to the intrinsic goals of the observer. These goals are driven by socially relevant stimuli addressed in our model through its reliance on multiple social cue modalities and facial information.
We detect the social and facial features in a separate stage, hereafter described as the \textit{Social Cue Detection} (SCD) stage.

To address the second finding, our model learns temporal features in two stages. Sequential learning in SCD is not a necessity but a result of the models employed for social cue detection, e.g., recurrent models or models pre-trained on optical flow tasks. In the second stage (GASP), we integrate social cues as illustrated in~\autoref{fig:full_model}. GASP also employs sequential learning, not only registering environmental changes such as color and intensity but also features pertaining to motion.  

Finally, we consider social attention by employing an audiovisual saliency prediction modality, as well as social cue detectors (also described as modalities) that specialize in performing distinct tasks. Each of these tasks is highly relevant to visual attention, both from a behavioral and a computational perspective. 
We aim to explore feature integration approaches for combining social cues. We present gated attention variants and introduce a novel approach for directing attention to all modalities. To the best of our knowledge, our model is the first to consider affect-biased attention by using facial expression representations for dynamic saliency prediction based on deep neural maps.

\section{Social Cue Detection}
\label{sec:stage1}
In the first stage (SCD), we extract high-level features from three social cue detectors and an audiovisual saliency predictor. We utilize the S$^{\text{3}}$FD face detector~\cite{zhang2017sfd} for acquiring the face locations of actors in an image. The cropped face images are passed to the social cue detectors as input. The window size $W$, i.e., the number of frames fed simultaneously as input to each model, varies according to the requirements of each model. We generate modality representations at output timestep $T'$ for each social video in AVE~\cite{tavakoli2020deep} as shown in \autoref{alg:stage1}.

\begin{algorithm}[h!]
\caption{SCD sampling and generation}
\label{alg:stage1}
\setstretch{0.8}
\textbf{Input}: \\
Video and audio frames sampled from $\texttt{ds}$ = AVE dataset \\
\textbf{Parameters}:\\
Window sizes $\texttt{W}_{\texttt{SP}}=15, \texttt{W}_{\texttt{GE}}=7, \texttt{W}_{\texttt{GF}}=5, \texttt{W}_{\texttt{FER}}=0$\\
O/P steps $\texttt{T'}_{\texttt{SP}}=15, \texttt{T'}_{\texttt{GE}}=4, \texttt{T'}_{\texttt{GF}}=0, \texttt{T'}_{\texttt{FER}}=0$\\
\textbf{Output}: \\
Modality windows $\texttt{mdl}_{\texttt{win}}$\\ 
O/P buffers $\texttt{buf}_{\texttt{mdl}}$
\begin{algorithmic}[1] 
\FOR{$\texttt{vid}$ in $\texttt{ds}$}
\STATE Let $\texttt{t}=0$
\FOR{$\texttt{frm}$ in $\texttt{vid}$}
\STATE Let $\texttt{fcs}$ = face crops \& bounding boxes in $\texttt{frm}$ 
\FOR{$\texttt{mdl}$ in $\in$ \{SP, GE, GF, FER\}}
\IF {$\texttt{W}_{\texttt{mdl}} > \texttt{t}$}
\STATE Let $\Delta = \texttt{W}_{\texttt{mdl}} - \texttt{t}$
\FOR{$\delta$ in $\in \{\Delta, ..., \texttt{W}_{\texttt{mdl}}\}$}
\STATE Let $\texttt{mdl}_{\texttt{win}}[\delta] =\:<\!\texttt{frm}, \texttt{fcs}\!>$
\ENDFOR
\ELSE
\STATE Shift $\texttt{mdl}_{\texttt{win}}$ by 1 to the left
\STATE Let $\texttt{mdl}_{\texttt{win}}[\texttt{W}_{\texttt{mdl}}] =\:<\!\texttt{frm},\texttt{fcs}\!>$
\ENDIF
\STATE Execute $\texttt{buf}_{\texttt{mdl}}[\texttt{t}] = \texttt{mdl}(\texttt{mdl}_{\texttt{win}})[\texttt{T'}_{\texttt{mdl}}]$
\ENDFOR
\STATE Propagate $\texttt{buf}[\texttt{t}]$ to GASP
\STATE Let $\texttt{t}=\texttt{t}+1$
\ENDFOR
\ENDFOR
\end{algorithmic}
\end{algorithm}

Our motivation behind selecting social cue detectors lies in their ability to generalize to various ``in-the-wild'' settings, regardless of the surrounding environment or lighting conditions. All chosen models were trained on datasets containing social entities captured from different angles and distances. 

\subsection{Gaze Following Model (GF)}

VideoGaze~\cite{recasens2017following} receives the source image frame containing the gazer, the target frame into which the gazer looks, and a face crop of the gazer in the source frame along with the head and eye positions as input to its pathways. 
All frames are resized to 227~$\times$~227 pixels. 
The model acquires five consecutive frames ($W_{GF}$) at timestep $T'_{GF}$ and returns a fixation heatmap of the most probable target frame for every detected face in a source frame. 

\textbf{Representation} The mean fixation heatmaps resulting from each face in the source frame are overlaid on a single feature map in the corresponding target frame timestep. We transform the fixation heatmaps using a jet colormap.

\subsection{Gaze Estimation Model (GE)}

Gaze360~\cite{kellnhofer2019gaze360} estimates the 3D gaze direction of an actor. The model receives face crops of the same actor over a predefined period, covering seven frames ($W_{GE}$) centered around timestep $T'_{GE}$. Each crop is resized to 224~$\times$~224 pixels. The model predicts the azimuth and pitch of the eyes and head along with a confidence score.

\textbf{Representation} 
We generate cones and position their tips on detected face centroids.
The cones are placed upon a zero-valued map with identical dimensions to the input image. The cone base is rotated towards the direction of gaze. 
The apex angle of the cone is set to $60^{\circ}$. The face furthest 
from the lens is projected first with an opacity of 0.5, followed by the remaining faces ordered by their distances to the lens. A jet colormap is then applied to the cone map.

\subsection{Facial Expression Recognition Model (FER)}

We employ the facial expression recognition model developed by Siqueira \textit{et al.}~\shortcite{siqueira2020efficient}. The model is composed of convolutional layers shared across $9$ ensembles.
The model receives all face crops in a frame as input, each resized to 96~$\times$~96 pixels, and recognizes facial expressions from 8 categories. Since the model operates on static images, we set the window size $W_{FER}$ and output timestep $T'_{FER}$ to 0.

\textbf{Representation}
Grad-CAM~\cite{selvaraju2017grad} features are extracted from all $9$ ensembles. We take the mean of the features for all faces in the image and apply a jet colormap transformation on them. A 2D Hanning filter is applied to the features to mitigate artifacts resulting from the edges of the cropped Grad-CAM representations. We center the filtered representations on the face positions upon a zero-valued map with dimensions identical to the input image.

\subsection{Audiovisual Saliency Prediction Model (SP)}

In the SCD stage, we utilize DAVE~\cite{tavakoli2020deep} for predicting saliency based on visual and auditory stimuli. Separate streams for encoding the two modalities are built using a 3D-ResNet with 18 layers. The visual stream acquires 16 images ($W_{SP}$), each resized to 256~$\times$~320 pixels. 
The auditory stream acquires log Mel-spectrograms of the video-corresponding audio frames, re-sampled to 16kHz. The model produces an FDM at the final output timestep $T'_{SP}$ considering all preceding frames within the window $W_{SP}$.

\textbf{Representation}
We transform the resulting FDM from DAVE using a jet colormap.

\section{Gated Attention for Saliency Prediction}
\label{sec:stage2}

We standardize all SCD features to a mean of 0 and a standard deviation of 1. The input image (IMG) and FMs are resized to 120~$\times$~120 pixels before propagation to GASP. 

\subsection{Directed Attention Module (DAM)}

The Squeeze-and-Excitation (SE)~\cite{hu2018squeeze} layer extracts channel-wise interactions, applying a gating mechanism for weighting convolutional channels according to their informative features. The SE model, however, emphasizes modality representations having the most significant gain, mitigating channels with lower information content. For our purpose, it is reasonable to postulate that the most influential FM channels are those belonging to the SP since it would result in the least erroneous representation in comparison to the ground-truth FDM. However, this causes the social cue modalities to have a minimal effect, mainly due to their low correlation with the FDM as opposed to the SP.

To counter bias towards the SP, we intensify non-salient regions such that the model learns to assign greater weights to modalities contributing least to the prediction. Alpay \textit{et al.}~\shortcite{alpay2019preserving} propose a language model for skipping and preserving activations according to how surprising a word is, given its context in a sequence.
In this work, we assimilate surprising words to channel regions with an unexpected contribution to the saliency model.

\begin{figure}[t]
\centering
\includegraphics[width = 0.5\textwidth]{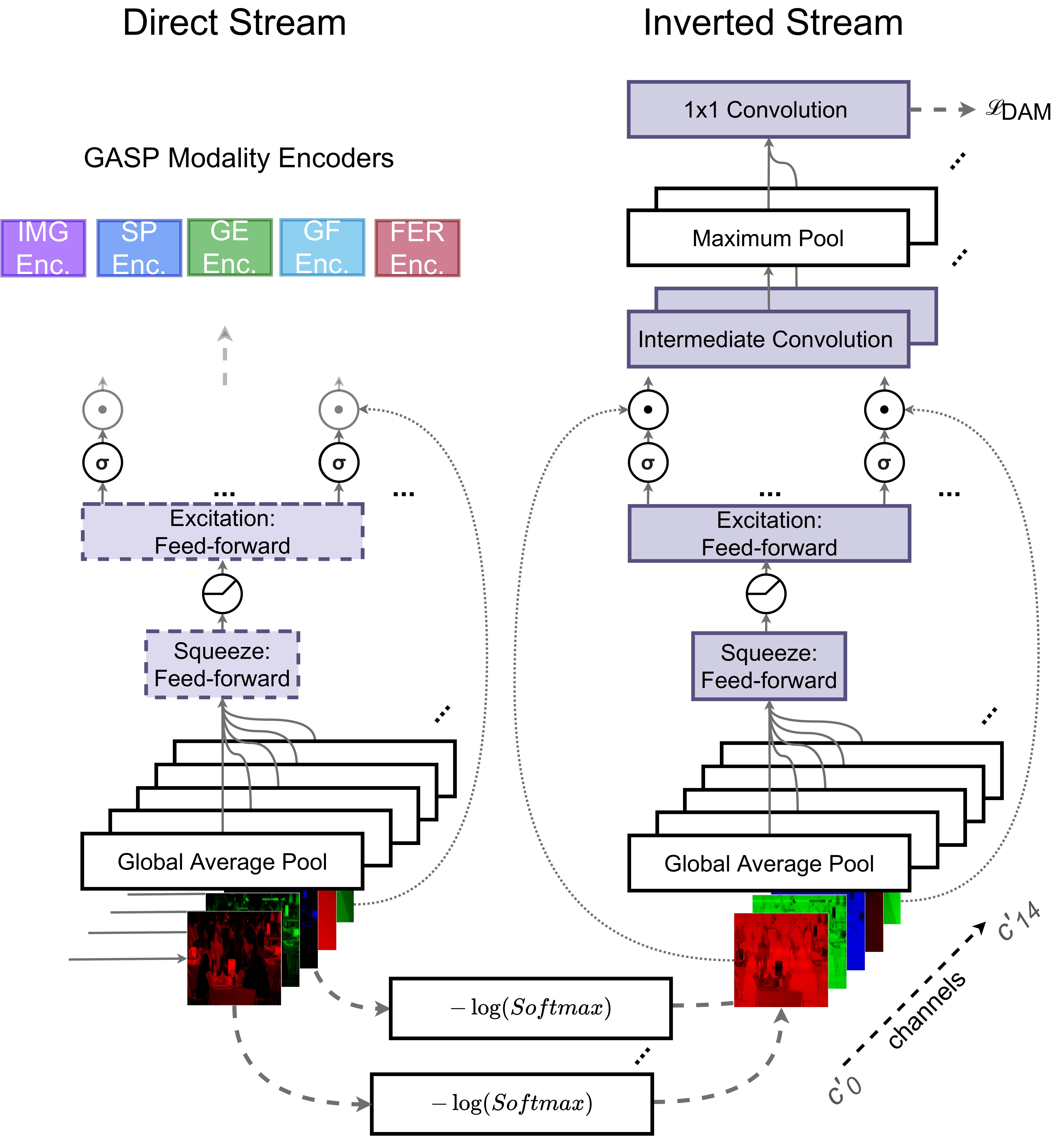}
\caption{The direct (left) and inverted (right) streams of our Directed Attention Module (DAM). The parameters of the direct stream are frozen and tied to the inverted stream as indicated by the dashed borders.}
\label{fig:dam}
\end{figure}

We construct a model for emphasizing unexpected features using two streams as shown in \autoref{fig:dam}: \textbf{1)} The inverted stream with output heads; \textbf{2)} The direct stream attached to the modality encoders of our GASP model. The inverted stream is composed of an SE layer followed by a 2D convolutional layer with a kernel size of 3~$\times$~3, a padding of 1, and 32 channels. A \textit{max pooling} layer with a window size of 2~$\times$~2 reduces the feature map dimensions by half. Finally, a 1~$\times$~1 convolution is applied to the pooled features, reducing the feature maps to a single channel. 
To emphasize weak features, we invert the input channels:
\begin{equation}
    \label{eq:surprisal}
    \mathbf{u}^{-1}_{c'} = \log \left( \frac{1}{Softmax(\mathbf{u}_{c'})} \right) = -\log[Softmax(\mathbf{u}_{c'})]
\end{equation}
where $\mathbf{u}_{c'}$ represents the individual channels of all modalities. The spatially inverted channels $\mathbf{u}^{-1}_{c'}$ are standardized and propagated as input features to the inverted stream. The direct stream is an SE layer with its parameters tied to the inverted stream and receives the standardized FM channels $\mathbf{u}_{c'}$ as input. Finally, the direct stream propagates the channel parameters multiplied with each FM to the modality encoders of GASP. The resulting weighted map is the priority map (PM).

\begin{figure*}[ht]
\centering

{\includegraphics[width = 0.475\textwidth]{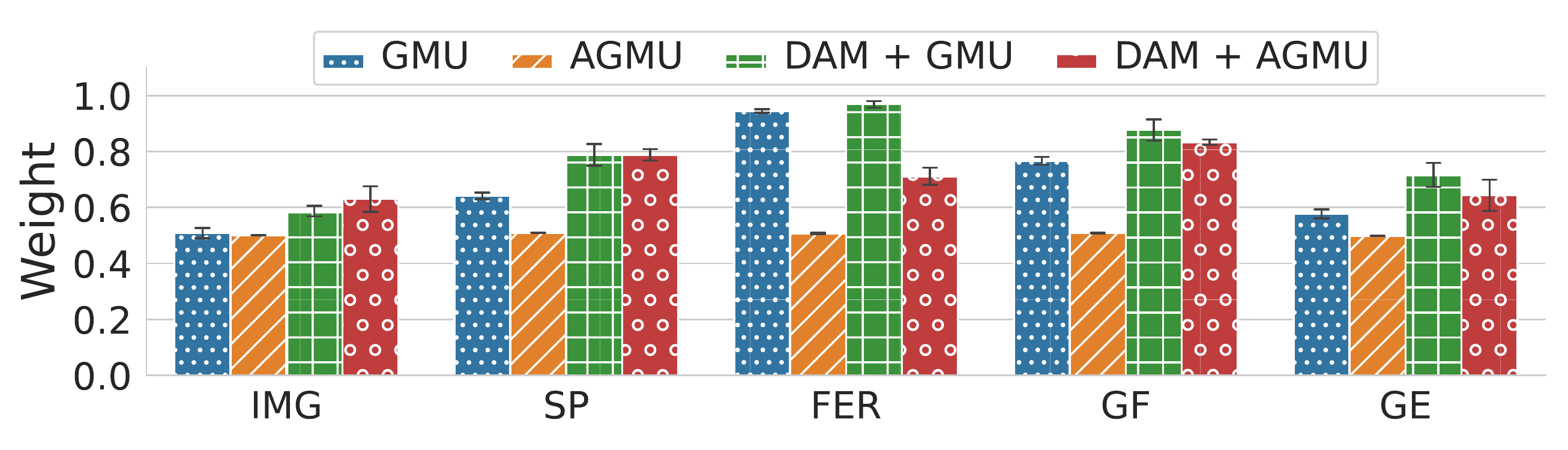}
\phantomsubcaption\label{cat}}
{\includegraphics[width = 0.495\textwidth]{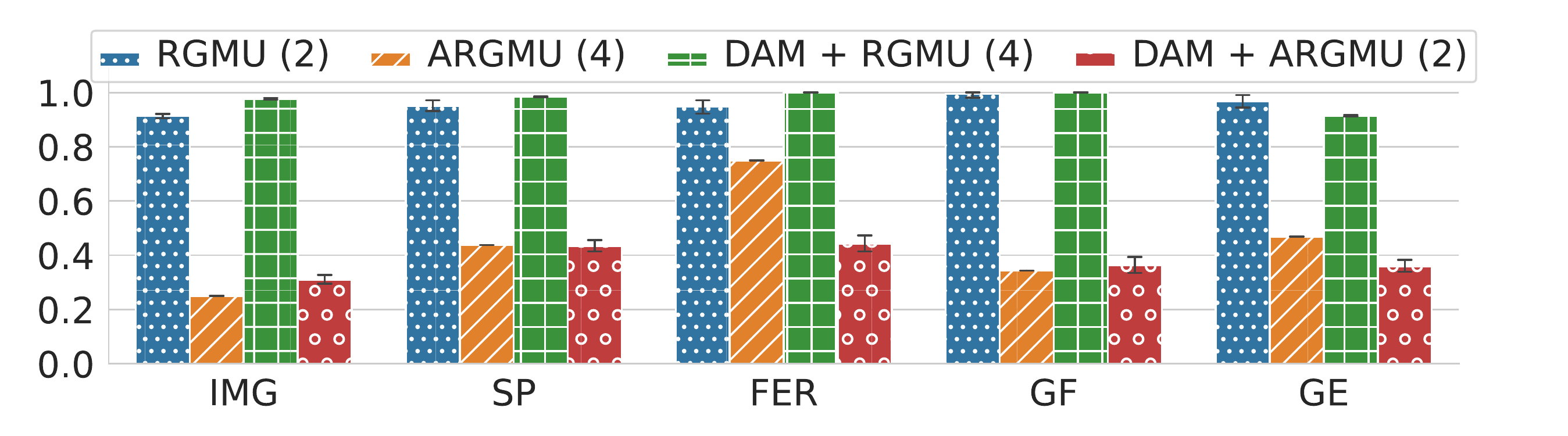}
\phantomsubcaption\label{bat}}



\caption{Aggregated modality weights of static (left) and sequential (right) fusion methods. Context sizes are shown within parentheses.}
\label{fig:combchart}
\end{figure*}

\subsection{Modality Encoder (Enc.)}
The modality encoder is a convolutional model used for extracting visual features from the priority maps. 
The first two layers of the encoder have 32 and 64 channels respectively. A maximum pooling layer reduces the input feature map to half its size. The pooled layer is followed by two layers with 128 channels each. Finally, the representations are decoded by applying transposed convolutions with 128, 64, and 32 channels. The last layer has a number of channels equivalent to the input channels. All convolutional kernels have a size of 3~$\times$~3, with a padding of 1. For GASP model variants operating on single frames (static integration variants), all modalities share the same encoder. For sequential integration variants, each modality has a separate encoder shared across timesteps.

\subsection{Recurrent Gated Multimodal Unit (RGMU)}

Concatenating the modality representations could lead to successful integration. Such a form of integration is commonly used in multimodal neural models, including audiovisual saliency predictors. We describe such approaches as non-fusion models, whereby the contribution of each modality is unknown. To account for all modalities, we employ the Gated Multimodal Unit (GMU)~\cite{arevalo2020gated}. The GMU learns to weigh the input features based on a gating mechanism.
To preserve the spatial features of the input, the authors introduce a convolutional variant of the GMU. This model, however, disregards the previous context since it does not integrate features sequentially. 
Therefore, we extend the convolutional GMU with recurrent units and express it as follows: 
\begin{equation}
    \label{eq:rgmu}
    \begin{array}{l}
    \mathbf{h}_{t}^{(m)} = \tanh{(\mathbf{W}_x^{(m)} * \mathbf{x}_{t}^{(m)} + \mathbf{U}_h^{(m)} * \mathbf{h}_{t-1}^{(m)} + b_h^{(m)})} \\
    \mathbf{z}_{t}^{(m)} = \sigma{(\mathbf{W}_z^{(m)} * [\mathbf{x}_{t}^{(1)}, ..., \mathbf{x}_{t}^{(M)}] + \mathbf{U}_z^{(m)} * \mathbf{z}_{t-1}^{(m)} + b_z^{(m)})} \\
    \mathbf{h}_{t} = \sum_{m=1}^{M} \mathbf{z}_t^{(m)} \odot \mathbf{h}_t^{(m)}
    \end{array}
\end{equation}
where $\mathbf{h}_{t}^{(m)}$ is the hidden representation of modality $m$ at timestep $t$. Similarly, $\mathbf{z}_{t}^{(m)}$ indicates the gated representation. The total number of modalities is represented by $M$. The parameters of the Recurrent Gated Multimodal Unit (RGMU) are denoted by $\mathbf{W}_x^{(m)}$, $\mathbf{W}_z^{(m)}$, $\mathbf{U}_h^{(m)}$, and $\mathbf{U}_z^{(m)}$. The modality inputs $\mathbf{x}^{(m)}$ at timestep $t$ are concatenated channel-wise as indicated by the $[\mathbf{\cdot}{,}\mathbf{\cdot}]$ operator and convolved with $\mathbf{W}_z^{(m)}$. The $\mathbf{z}_t^{(m)}$ representation is acquired by summing the current and previous timestep representations, along with the bias term $b_z^{(m)}$. A sigmoid activation function denoted by $\sigma$ is applied to the recurrent representations $\mathbf{z}_t$. The final feature map $\mathbf{h}_t$ is the Hadamard-product between $\mathbf{z}_t^{(m)}$ and $\mathbf{h}_t^{(m)}$ summed over all modalities. 

The aforementioned recurrent approach suffers from vanishing gradients as the context becomes longer. To remedy this effect, we propose the integration of GMU with the convolutional Attentive Long Short-Term Memory (ALSTM)~\cite{cornia2018predicting}. 
ALSTM applies soft-attention to single timestep input features over multiple iterations.
We utilize ALSTM for our static GASP integration variants. For sequential variants, we  modify ALSTM to acquire frames at all timesteps instead of attending to a single frame multiple times:
\begin{equation}
    \label{eq:alstm_atten}
    \mathbf{x}_t = Softmax(\mathbf{z}_{t-1}) \odot \mathbf{x}_t
\end{equation}
where $\mathbf{z}_{t-1}$ represents the pre-attentive output of the previous timestep. We adapt the sequential ALSTM to operate in conjunction with the GMU by performing the gated fusion per timestep. We refer to this model as the Attentive Recurrent Gated Multimodal Unit (ARGMU). Alternatively, we perform the gated integration after concatenating the input channels and propagating them to the sequential ALSTM. Since the modality representations are no longer separable, we describe this variant as the Late ARGMU (LARGMU). We refer to the total number of timesteps as the context size. Analogous to the sequential variants, we create similar gating mechanisms for static integration approaches. Replacing the sequential ALSTM with the ALSTM by Cornia \textit{et al.}~\shortcite{cornia2018predicting}, we present the non-sequential Attentive Gated Multimodal Unit (AGMU), as well as the Late AGMU (LAGMU).

\section{Experimental Setup}
\label{sec:exp}

We train our GASP model on the social event subset of AVE~\cite{tavakoli2020deep}. AVE is a composition of three datasets: DIEM~\cite{mital2011clustering}, Coutrot Databases 1~\cite{coutrot2014audiovisual} and 2~\cite{coutrot2015efficient}.
To train the model, we employ the loss functions introduced by Tsiami \textit{et al.}~\shortcite{tsiami2020stavis}, assigning the loss weights $\lambda_1=0.1$, $\lambda_2=2$, and $\lambda_3=1$ to \textit{cross-entropy}, \textit{CC}, and \textit{NSS} losses respectively. The loss functions $ \mathcal{L}_{\!P\!F\!D\!M}$ are weighted, summed, and applied to the final layer for optimizing the modality encoder and integration model parameters. 
The model is trained using the Adam optimizer, having a learning rate of 0.001, with $\beta_1=0.9$ and $\beta_2=0.999$. 
All models are trained for $\sim$10k iterations with a batch size of 4. We conduct five trials acquiring the mean results for evaluation.

The models are evaluated on the test subset of social event videos in AVE.
We employ five commonly used metrics in dynamic saliency prediction~\cite{tavakoli2020deep,tsiami2020stavis}: 
\textit{Normalized scanpath saliency} (NSS); \textit{Linear correlation coefficient} (CC); \textit{Similarity metric} (SIM); \textit{Area under the ROC curve} (AUC-J); \textit{Shuffled AUC} (sAUC). The negative fixations for the sAUC metric are sampled from all the mean eye positions in the social event subset of AVE.

The inverted stream of our DAM layer has a separate output head for each timestep. We compute the cross-entropy between the DAM prediction and the FDM. For sequential integration models, the loss is summed over all timesteps. The loss $\mathcal{L}_{\!D\!A\!M}$ with a weight $\lambda_{\!D\!A\!M}=0.5$ is computed for optimizing the inverted stream parameters. The parameters are transferred to the direct stream with frozen parameters.

An NVIDIA RTX 2080 Ti GPU with 11 GB VRAM and 128 GB RAM is used for training all static and sequential models. To extract spatiotemporal maps in the first stage (SCD), we employ an NVIDIA TITAN RTX GPU with 24 GB VRAM and 64 GB RAM to accommodate all social cue detectors simultaneously. We perform the SCD feature extraction in a preprocessing step for all videos in the AVE dataset. 
 
\section{Results}
\label{sec:results}

\subsection{Static Integration}

\begin{table}[H]
\centering
\caption{Static integration results. Top rows represent non-fusion methods and bottom rows are fusion-based integration approaches.}
\label{tab:feedforward_res}
\resizebox{\columnwidth}{!}{%
\setlength\tabcolsep{4pt}
\begin{tabular}{c|ccccc}
 \hline
 \textbf{Model} & AUC-J $\uparrow$ & sAUC $\uparrow$ & CC $\uparrow$ & NSS $\uparrow$ & SIM $\uparrow$ \\ \hline
 Additive & 0.5842 & 0.5912 & 0.0882 & 1.19 & 0.1878 \\ \hline 
 Concatenative & 0.8782 & 0.6303 & 0.6614 & 2.71 & 0.4743 \\ \hline 
 ALSTM & 0.6881 & 0.5727 & 0.4503 & 2.05 & 0.3316 \\ \hline 
 SE & 0.5367 & 0.5597 & 0.0359 & 1.03 & 0.0972 \\ \hline 
 LAGMU \textit{(Ours)} & 0.8347 & 0.6376 & 0.5576 & 2.48 & 0.4361 \\ \hline 
 DAM + LAGMU \textit{(Ours)} & 0.8791 & 0.6379 & 0.6606 & 2.76 & \textbf{0.5278} \\ \hline \hline 
 GMU & 0.8792 & 0.6374 & 0.6545 & 2.75 & 0.5172 \\ \hline 
 AGMU \textit{(Ours)} & 0.6829 & 0.6359 & 0.2046 & 1.47 & 0.2212 \\ \hline 
 DAM + GMU \textit{(Ours)} & \textbf{0.8845} & \textbf{0.6397} & \textbf{0.6620} & \textbf{2.77} & 0.5233 \\ \hline 
 DAM + AGMU \textit{(Ours)} & 0.8587 & 0.6372 & 0.6372 & 2.71 & 0.5066 \\ \hline 
 
 \end{tabular}%
 }
\end{table}

We examine integration approaches operating on a single frame in GASP. The \textit{Additive} model refers to the integration variant in which the feature maps of all encoders are summed, followed by a 3~$\times$~3 convolution with 32 channels and a padding of 1. The \textit{Concatenative} variant applies a channel-wise concatenation to the feature maps, followed by the aforementioned convolutional layer. ALSTM, LAGMU, and AGMU employ the non-sequential ALSTM variant by Cornia \textit{et al.}~\shortcite{cornia2018predicting}. The Squeeze-and-Excitation~\cite{hu2018squeeze} (SE) model precedes the modality encoder. We note that all models excluding SE and DAM replace the integration model. Finally, all model variants are followed by a 1~$\times$~1 convolution resulting in the final output feature map.

In~\autoref{tab:feedforward_res}, we show that the best NSS scores are achieved by our DAM + GMU variant. This indicates fewer false positives predicted by fusion in comparison to other non-fusion methods. We also observe that the DAM enhances fusion models but appears to have minimal effect on the GMU variant (the difference across all metrics is insignificant).

\subsection{Sequential Integration}

\begin{table*}[htbp!]
\centering
\caption{Sequential integration results. Top rows represent non-fusion methods and bottom rows are fusion-based integration approaches.}
\label{tab:recurrent_res}
\setlength\tabcolsep{1.5pt}
\begin{tabular}{c||c|c|c}

\begin{subtable}[h]{0.175\textwidth}
    \resizebox{\textwidth}{!}{%
    \begin{tabular}{c}
     \hline
     \textbf{Model}\\ \hline
     Sequential ALSTM\\ \hline
     LARGMU \textit{(Ours)}\\ \hline 
     DAM + LARGMU \textit{(Ours)}\\ \hline \hline 
     RGMU \textit{(Ours)}\\ \hline 
     ARGMU \textit{(Ours)}\\ \hline 
     DAM + RGMU \textit{(Ours)}\\ \hline 
     DAM + ARGMU \textit{(Ours)}\\ \hline 
     \end{tabular}%
     }
     \caption{}
    \label{tab:recurrent_res_r1}
\end{subtable} &

\begin{subtable}[h]{0.26\textwidth}
    \resizebox{\textwidth}{!}{%
    \begin{tabular}{ccccc}
     \hline
     AUC-J $\uparrow$ & sAUC $\uparrow$ & CC $\uparrow$ & NSS $\uparrow$ & SIM $\uparrow$ \\ \hline
     
     \textbf{0.8849} & 0.6428 & 0.6590 & 2.79 & 0.4522\\ \hline 
     0.8791 & \textbf{0.6444} & 0.6572 & 2.76 & 0.5251\\ \hline 
     0.8789 & 0.6437 & \textbf{0.6703} & \textbf{2.78} & 0.5354\\ \hline \hline 
     0.8343 & 0.6239 & 0.6195 & 2.62 & 0.4717\\ \hline 
     0.8793 & 0.6410 & 0.6607 & 2.74 & \textbf{0.5359}\\ \hline 
     0.8726 & 0.6329 & 0.6547 & 2.73 & 0.5135\\ \hline 
     0.8747 & 0.6421 & 0.6536 & \textbf{2.78} & 0.5305\\ \hline 
     \end{tabular}%
     }
    \caption{Context Size = 2}
    \label{tab:recurrent_res2}
\end{subtable} &
\begin{subtable}[h]{0.26\textwidth}
    \resizebox{\textwidth}{!}{%
    \begin{tabular}{ccccc}
     \hline
     AUC-J $\uparrow$ & sAUC $\uparrow$ & CC $\uparrow$ & NSS $\uparrow$ & SIM $\uparrow$ \\ \hline
     
     0.8794 & 0.6439 & 0.6621 & 2.78 & 0.5244\\ \hline 
     \textbf{0.8860} & \textbf{0.6460} & \textbf{0.6698} & \textbf{2.80} & 0.5191\\ \hline 
     0.8799 & 0.6443 & 0.6568 & 2.76 & \textbf{0.5287}\\ \hline \hline 
     0.8797 & 0.6350 & 0.6184 & 2.69 & 0.4614\\ \hline 
     0.8819 & 0.6456 & 0.6556 & 2.75 & 0.5279\\ \hline 
     0.8714 & 0.6345 & 0.6539 & 2.73 & 0.5198\\ \hline 
     0.8790 & 0.6363 & 0.6391 & 2.72 & 0.4934\\ \hline 
     \end{tabular}%
     }
    \caption{Context Size = 4}
    \label{tab:recurrent_res4}
\end{subtable} &

\begin{subtable}[h]{0.26\textwidth}
    \resizebox{\textwidth}{!}{%
    \begin{tabular}{ccccc}
     \hline
     AUC-J $\uparrow$ & sAUC $\uparrow$ & CC $\uparrow$ & NSS $\uparrow$ & SIM $\uparrow$ \\ \hline
     
     0.8789 & 0.6425 & \textbf{0.6612} & \textbf{2.76} & 0.5291\\ \hline 
     \textbf{0.8818} & \textbf{0.6433} & 0.6556 & \textbf{2.76} & 0.5205\\ \hline 
     0.8801 & 0.6423 & 0.6589 & \textbf{2.76} & \textbf{0.5293}\\ \hline \hline 
     0.7331 & 0.5851 & 0.5117 & 2.33 & 0.3823\\ \hline 
     0.8656 & 0.6388 & 0.6534 & 2.72 & 0.5017\\ \hline 
     0.8718 & 0.6346 & 0.6510 & 2.73 & 0.5172\\ \hline 
     0.8560 & 0.6271 & 0.6418 & 2.70 & 0.5133\\ \hline 
     \end{tabular}%
     }
    \caption{Context Size = 6}
    \label{tab:recurrent_res6}
\end{subtable} \\

\begin{subtable}[h]{0.18\textwidth}
    \resizebox{\textwidth}{!}{%
    \begin{tabular}{c}
     \hline
     \textbf{Model}\\ \hline
     Sequential ALSTM\\ \hline
     LARGMU \textit{(Ours)}\\ \hline 
     \textbf{DAM + LARGMU \textit{(Ours)}}\\ \hline \hline 
     RGMU \textit{(Ours)}\\ \hline 
     ARGMU \textit{(Ours)}\\ \hline 
     DAM + RGMU \textit{(Ours)}\\ \hline
     DAM + ARGMU \textit{(Ours)}\\ \hline 
     \end{tabular}%
     }
     \caption{}
    \label{tab:recurrent_res_r2}
\end{subtable} &

\begin{subtable}[h]{0.26\textwidth}
    \resizebox{\textwidth}{!}{%
    \begin{tabular}{ccccc}
     \hline
     AUC-J $\uparrow$ & sAUC $\uparrow$ & CC $\uparrow$ & NSS $\uparrow$ & SIM $\uparrow$ \\ \hline
     
     0.8766 & 0.6389 & 0.6628 & 2.75 & 0.5307\\ \hline 
     0.8702 & 0.6362 & 0.6529 & 2.72 & 0.5307\\ \hline 
     \textbf{0.8872} & \textbf{0.6529} & \textbf{0.6903} & \textbf{2.84} & \textbf{0.5520}\\ \hline \hline 
     0.7982 & 0.6031 & 0.5763 & 2.48 & 0.4368\\ \hline 
     0.6892 & 0.5680 & 0.3253 & 1.84 & 0.2549\\ \hline 
     0.8678 & 0.6344 & 0.6622 & 2.75 & 0.5212\\ \hline 
     0.8580 & 0.6259 & 0.6274 & 2.66 & 0.4874\\ \hline 

     \end{tabular}%
     }
    \caption{Context Size = 8}
    \label{tab:recurrent_res8}
\end{subtable} &

\begin{subtable}[h]{0.26\textwidth}
    \resizebox{\textwidth}{!}{%
    \begin{tabular}{ccccc}
     \hline
     AUC-J $\uparrow$ & sAUC $\uparrow$ & CC $\uparrow$ & NSS $\uparrow$ & SIM $\uparrow$ \\ \hline
     
     0.8773 & 0.6412 & 0.6704 & 2.76 & 0.5306\\ \hline 
     0.8791 & 0.6356 & 0.6511 & 2.72 & 0.5168\\ \hline 
     \textbf{0.8830} & \textbf{0.6527} & \textbf{0.6980} & \textbf{2.87} & \textbf{0.5566} \\ \hline \hline 
     0.8229 & 0.5981 & 0.5197 & 2.53 & 0.4502\\ \hline 
     0.8467 & 0.6179 & 0.6116 & 2.62 & 0.4707\\ \hline 
     0.8579 & 0.6284 & 0.6540 & 2.72 & 0.5207\\ \hline 
     0.8663 & 0.6303 & 0.6446 & 2.69 & 0.5161\\ \hline 

     \end{tabular}%
     }
    \caption{\textbf{Context Size = 10}}
    \label{tab:recurrent_res10}
\end{subtable} &

\begin{subtable}[h]{0.26\textwidth}
    \resizebox{\textwidth}{!}{%
    \begin{tabular}{ccccc}
     \hline
     AUC-J $\uparrow$ & sAUC $\uparrow$ & CC $\uparrow$ & NSS $\uparrow$ & SIM $\uparrow$ \\ \hline
     
     0.8759 & 0.6352 & \textbf{0.6665} & 2.74 & 0.5275\\ \hline 
     \textbf{0.8788} & 0.6416 & 0.6624 & \textbf{2.75} & 0.5152\\ \hline 
     0.8775 & \textbf{0.6418} & 0.6612 & 2.74 & \textbf{0.5328}\\ \hline \hline 
     0.8147 & 0.5717 & 0.4119 & 2.30 & 0.3276\\ \hline 
     0.8457 & 0.6130 & 0.6007 & 2.56 & 0.4593\\ \hline 
     0.8759 & 0.6327 & 0.6549 & 2.74 & 0.5123\\ \hline 
     0.8561 & 0.6249 & 0.6441 & 2.70 & 0.5117\\ \hline 

     \end{tabular}%
     }
    \caption{Context Size = 12}
    \label{tab:recurrent_res12}
\end{subtable} 

\end{tabular}
 \end{table*}
 
We modify our GASP integration model to have a context greater than one. 
All models employ batch normalization applied to the temporal axis. The integration models are followed by a 1~$\times$~1 convolution resulting in the final output feature map. In \autoref{tab:recurrent_res}, we experiment with context sizes $\in\{2,4,6,8,10,12\}$ and observe an overall improvement in performance with a context size of 4. The directed attention variant with late non-fusion gating (DAM + LARGMU) achieves the best scores on all metrics. This implies that gated integration is beneficial, even though the representations preceding the GMU are not separable. 

Comparing the results of static integration in~\autoref{tab:feedforward_res} to dynamic integration approaches in~\autoref{tab:recurrent_res}, we observe that several static approaches perform on a par with recurrent models. Nonetheless, the sequential DAM + LARGMU with context sizes of 8 and 10 outperform all integration methods. In~\autoref{tab:detailed_best_res}, we observe an insignificant difference in metric scores among the best sequential models for all context sizes. Compared to the best static model, the variances of sequential model scores are lower, indicating the stabilizing influence of attentive LSTMs with the addition of context.

\subsection{Modality Contribution}

\begin{table}[H]
\centering
\caption{Social cue modality ablation applied to our best GASP model (DAM + LARGMU; Context Size = 10).}
\label{tab:ablation_res}
\resizebox{\columnwidth}{!}{%
\setlength\tabcolsep{6pt}
\begin{tabular}{c|c|c|ccccc}
 \hline
 \textbf{GE} & \textbf{GF} & \textbf{FER} & AUC-J $\uparrow$ & sAUC $\uparrow$ & CC $\uparrow$ & NSS $\uparrow$ & SIM $\uparrow$ \\ \hline

 - & - & - & 0.8767 & 0.6338 & 0.6542 & 2.72 & 0.5228 \\ \hline
 - & - & \checkmark & 0.7535 & 0.5951 & 0.4466 & 2.17 & 0.3578 \\ \hline
 - & \checkmark & - & 0.6893 & 0.5679 & 0.3222 & 1.84 & 0.2539 \\ \hline
 - & \checkmark & \checkmark & 0.8778 & 0.6442 & 0.6652 & 2.76 & 0.5350 \\ \hline
 \checkmark & - & - & 0.8769 & 0.6272 & 0.6493 & 2.70 & 0.4798 \\ \hline
 \checkmark & - & \checkmark & \textbf{0.8859} & 0.6505 & 0.6840 & 2.86 & 0.5381 \\ \hline
 \checkmark & \checkmark & - & 0.8776 & 0.6367 & 0.6543 & 2.74 & 0.5216 \\ \hline
 \checkmark & \checkmark & \checkmark & 0.8830 & \textbf{0.6527} & \textbf{0.6980} & \textbf{2.87} & \textbf{0.5566} \\ \hline
 \end{tabular}%
 }
 \end{table}
 
We evaluate the contribution of each modality to the final prediction by computing the mean activation of the gates across channels and timesteps. This evaluation is only applicable to fusion methods in either static or sequential forms of integration. We observe that the DAM does not alter the modality contribution of the static GMU. For sequential variants, introducing the DAM allows modalities to have a uniform contribution to the final output.

We examine the modalities contributing an improvement to the best non-fusion sequential model. As shown in~\autoref{tab:ablation_res}, FER in combination with GE achieves results on par with the best model. The exclusion of GF has a minimal effect on the model due to the sparsity of its representation. Significant degradation in the model variant with social modalities ablated implies the necessity of social cues in concert.

\subsection{Comparison with State-of-the-art}

\begin{table}[H]
\centering
\caption{Comparison with state-of-the-art by varying the SCD SP of our best GASP model (DAM + LARGMU; Context Size = 10).}
\label{tab:sota_res}
\resizebox{\columnwidth}{!}{%
\setlength\tabcolsep{1pt}
\begin{tabular}{c|c|ccccc}
 \hline
 
 \textbf{Model} & Test & AUC-J $\uparrow$ & sAUC $\uparrow$ & CC $\uparrow$ & NSS $\uparrow$ & SIM $\uparrow$ \\ \hline
 
 UNISAL (Visual only) & AVE & 0.8640 & 0.6545 & 0.4243 & 2.04 & 0.3818 \\ \hline 
 TASED (Visual only) & AVE & 0.8601 & 0.6515 & 0.4631 & 2.19 & 0.4084 \\ \hline 
 DAVE (Visual only) & AVE & 0.8824 & 0.6138 & 0.5136 & 2.45 & 0.4080 \\ \hline 
 DAVE (Audiovisual baseline) & AVE & 0.8853 & 0.6121 & 0.5453 & 2.65 & 0.4420 \\ \hline 
 STAViS (Visual only) & STA & 0.8577 & 0.6517 & 0.4690 & 2.08 & 0.4004 \\ \hline 
 STAViS (Audiovisual) & STA & 0.8752 & 0.6154 & 0.4912 & 2.79 & 0.4774 \\ \hline 
  \hline
 UNISAL + GASP \textit{(Ours)} & AVE & 0.8771 & 0.6334 & 0.6494 & 2.70 & 0.5244 \\ \hline 
 TASED + GASP \textit{(Ours)} & AVE & 0.8602 & 0.6195 & 0.5736 & 2.50 & 0.4725 \\ \hline 
 DAVE + GASP \textit{(Ours)} & AVE & 0.8830 & 0.6527 & \textbf{0.6980} & 2.87 & \textbf{0.5566} \\ \hline 
 STAViS + GASP \textit{(Ours)} & STA & \textbf{0.8910} & \textbf{0.6825} & 0.6052 & \textbf{3.08} &  0.4324 \\ \hline 

 \end{tabular}%
 }
 \end{table}
 
We compare the performance of our model with four dynamic saliency predictors. We replace DAVE~\cite{tavakoli2020deep} with STAViS~\cite{tsiami2020stavis}, TASED~\cite{min2019tased} and UNISAL~\cite{droste2020unifiedia} in the SCD stage during the evaluation phase. 
Due to the overlap in datasets between DAVE and STAViS, we retrain our GASP model with STAViS as the SCD audiovisual saliency predictor. We evaluate and train our STAViS-based model on social event videos according to the data splits concocted by Tsiami \textit{et al.}~\shortcite{tsiami2020stavis}. 

\begin{figure*}[hbt!]
\centering

\setlength{\tabcolsep}{1pt} 
\renewcommand{\arraystretch}{0.5} 
\includegraphics[width = 1\textwidth]{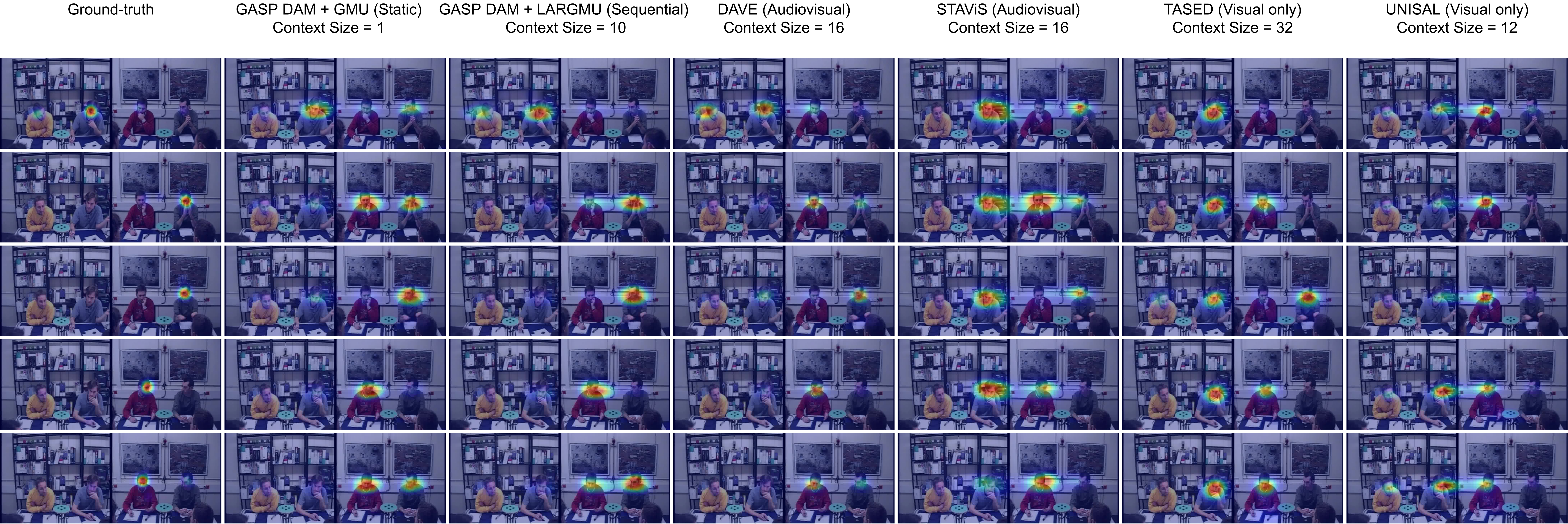}
\caption{Frame predictions on the Coutrot Database 2~\protect\cite{coutrot2015efficient} for comparison with state-of-the-art models. Our GASP models employ the audiovisual DAVE model (Context Size = 16) as the saliency predictor in the SCD stage.}
\label{fig:sota_samples}
\end{figure*}

\begin{table*}[hbpt!]
\centering
\caption{Best model results for all context sizes with the standard deviation for each metric score provided, along with the training duration and number of parameters. All models employ the audiovisual DAVE model (Context Size = 16) as the saliency predictor in the SCD stage.}
\label{tab:detailed_best_res}
\resizebox{\textwidth}{!}{%
\setlength\tabcolsep{5pt}
\begin{tabular}{c|ccccc|ccc}
 \hline
 \textbf{Model} & AUC-J $\uparrow$ & sAUC $\uparrow$ & CC $\uparrow$ & NSS $\uparrow$ & SIM $\uparrow$ & No. Parameters & Training Time & Context Size\\ \hline
 
 DAM + GMU & 0.8845 $\pm$ 0.1389 & 0.6397 $\pm$ 0.0831 & 0.6620 $\pm$ 0.2324 & 2.77 $\pm$ 0.53 & 0.5233 $\pm$ 0.1628 & 0.77M & 25 mins & 1\\ \hline
 DAM + LARGMU & 0.8789 $\pm$ 0.0420 & 0.6437 $\pm$ 0.0742 & 0.6703 $\pm$ 0.1068 & 2.78 $\pm$ 0.32 & 0.5354 $\pm$ 0.0675  & 4.28M & 185 mins & 2\\ \hline
 LARGMU & 0.8860 $\pm$ 0.0396 & 0.6460 $\pm$ 0.0778 & 0.6698 $\pm$ 0.1093 & 2.80 $\pm$ 0.32 & 0.5191 $\pm$ 0.0634 & 4.28M & 140 mins & 4\\ \hline
 LARGMU & 0.8818 $\pm$ 0.0400 & 0.6433 $\pm$ 0.0732 & 0.6556 $\pm$ 0.1027 & 2.76 $\pm$ 0.30 & 0.5205 $\pm$ 0.0607 & 4.28M & 130 mins & 6\\ \hline
 DAM + LARGMU & 0.8872 $\pm$ 0.0374 & 0.6529 $\pm$ 0.0703 & 0.6903 $\pm$ 0.1167 & 2.84 $\pm$ 0.31 & 0.5520 $\pm$ 0.0729 & 4.28M & 280 mins & 8\\ \hline
 DAM + LARGMU & 0.8830 $\pm$ 0.0451 & 0.6527 $\pm$ 0.0785 & 0.6980 $\pm$ 0.1164 & 2.87 $\pm$ 0.34 & 0.5566 $\pm$ 0.0753 & 4.28M & 315 mins & 10\\ \hline
 DAM + LARGMU & 0.8775 $\pm$ 0.0430 & 0.6418 $\pm$ 0.0747 & 0.6612 $\pm$ 0.1115 & 2.74 $\pm$ 0.32 & 0.5328 $\pm$ 0.0716 & 4.28M & 330 mins & 12\\
 \hline
 \end{tabular}%
 }
\end{table*}

Combining our best GASP model with different saliency predictors improves their performances, as shown in~\autoref{tab:sota_res}. Although the GASP model is not retrained, it extracts information from the social cue modalities and the saliency predictor (SP) pertinent to the prediction. The sequential GASP also exhibits greater resistance to central bias as shown in~\autoref{fig:sota_samples} (middle row) compared to other models, where the actor closest to the center is incorrectly predicted as a fixation target. The integration of social cue features and sequential inference in both stages of GASP contribute to such resistance.

\section{Conclusion}
\label{sec:conc}

We introduce GASP, a gated attention model for predicting saliency by integrating knowledge from two social cues using three separate detectors. We represent each social cue as a spatiotemporal map, accounting for the gaze direction, gaze target, and facial expressions of an observed individual. We examine gated and recurrent approaches for integrating the social cues. Our gated integration variants achieve better results than non-gated approaches. We also present a module for emphasizing weak features, which is effective for static and sequential integration methods. Furthermore, we show that gaze direction and facial expression representations have a positive effect when integrated with saliency models. The latter supports the importance of considering affect-biased attention. Coupled with state-of-the-art saliency models, GASP improves their prediction performances on multiple metrics, indicating the efficacy of social cue integration in our architecture.




\section*{Acknowledgements}
The authors gratefully acknowledge partial support from the German
Research Foundation DFG under project CML ~(TRR 169).







\bibliographystyle{named}
\bibliography{ijcai21}

\end{document}